# A Marine Debris Detection Framework for Ocean Robots via Self-Attention Enhancement and Feature Interaction Optimization

Yuyang Li, Jiashu Han, Yinyi Lai, Wenbin Kang, and Zenghui Liu

***Abstract*—Marine debris detection for ocean robot is crucial for ecological protection, yet performance is often degraded by low-quality images with blur, complex backgrounds, and small targets. To address these challenges, we propose YOLO-MD, an enhanced YOLO-based detection framework. A Dual-Branch Convolutional Enhanced Self-Attention (DB-CASA) module is designed to strengthen spatial–channel interactions, improving feature representation in degraded images. Additionally, a lightweight shift-based operation is introduced to enhance fine-grained feature extraction for objects of varying scales while maintaining parameter efficiency. We further propose SFG-Loss to mitigate class imbalance and optimization instability via dynamic sample reweighting. Experiments on the UODM dataset demonstrate that YOLO-MD achieves 0.875 precision, 0.822 F1-score, and 0.849 mAP50, outperforming latest state-of-the-art methods The effectiveness of this method has also been verified through real-world robotic edge deployment experiments.**

***Index Terms*—Marine debris detection, Ocean robot, YOLO, Self-attention, Shift channel, Class imbalance.**

## I. Introduction

The problem of marine debris poses a significant threat to ecosystems and human health [1]. Accurate detection and localization is essential for effective pollution control. However, marine images are often degraded by blur, complex backgrounds, and small, densely distributed objects, which significantly hinder detection performance for ocean robots[2]. Traditional methods improve image quality through enhancement, deblurring, and super-resolution. However, these approaches introduce substantial computational overhead and are unsuitable for real-time applications [3] [4] [5].

The work was supported in part by the Natural Science Foundation of Jiangsu Province, China (BK20241781), the Fundamental Research Funds for the Central Universities (B250203012), and the National Natural Science Foundation (52371275). (*Corresponding author: Yinyi Lai, Wenbin Kang, and Zenghui Liu.*)

Yuyang Li and Zenghui Liu are with the School of Department of Mechanical Engineering, Hohai University, Changzhou, 213200, China (e-mail: forward0412@163.com; zenghui.liu@hhu.edu.cn)

Jiashu Han is with the Hefei Institutes of Physical Science, Chinese Academy of Sciences, Hefei, 230031, China (e-mail: 610360177@qq.com)

Yinyi Lai and Wenbin Kang are with the Department of Mechanical Engineering, City University of Hong Kong, Hong Kong SAR, 999077, China (e-mail: yinyilai2-c@my.cityu.edu.hk; wenbin.kang@cityu.edu.hk)

Yuyang Li, Jiashu Han, and Yinyi Lai contributed equally to this work.

Existing object detection methods can be broadly categorized into three types. Two-stage detectors, such as Faster R-CNN [6], achieve high accuracy via region proposals but suffer from high computational cost and slow inference. One-stage detectors, including SSD [7] and the YOLO series [8], enable efficient end-to-end detection and are more suitable for real-time scenarios. Among them, YOLO achieves a favorable balance between speed and accuracy, yet its performance degrades in blurred scenes and small-object detection. In addition, there are also some methods based on the YOLO series algorithms for improvement, such as MAS-YOLOv11 [9], CEH-YOLO [10] and DMFI-YOLO [11], which mark the continuous innovation of object detection algorithms.

Recently, Transformer-based detectors, such as RT-DETR[12], provide strong global modeling capability. However, their reliance on global attention limits sensitivity to local fine-grained features, and their computational complexity restricts deployment in real-time and resource-constrained scenarios.

Despite these advances, effectively capturing fine-grained features of small targets and addressing class imbalance in complex marine environments remain challenging [13].

To address these issues, we propose YOLO-MD, a YOLO-based marine debris detection framework with enhanced feature representation and detection performance. The main contributions are summarized as follows:

- We replace the C2PSA module in the YOLO backbone with a Dual-Branch Convolutional Enhanced Self-Attention module, which uses an additive similarity function and parallel spatial-channel interactions to preserve fine-grained features of small targets in low-quality images.
- To better capture fine-grained features of small targets, we design a Feature Shift Fusion Module, which employs a cross-dimensional feature exchange strategy via grouped spatial shifts to enhance small-object detection precision without introducing additional parameters.
- We propose SFG-Loss to mitigate class imbalance and optimization instability via dynamic sample reweighting.

Extensive experiments on the UODM dataset demonstrate that YOLO-MD outperforms representative detectors across multiple metrics. In addition, real-world deployment experiments further validate its detection performance and practical applicability.

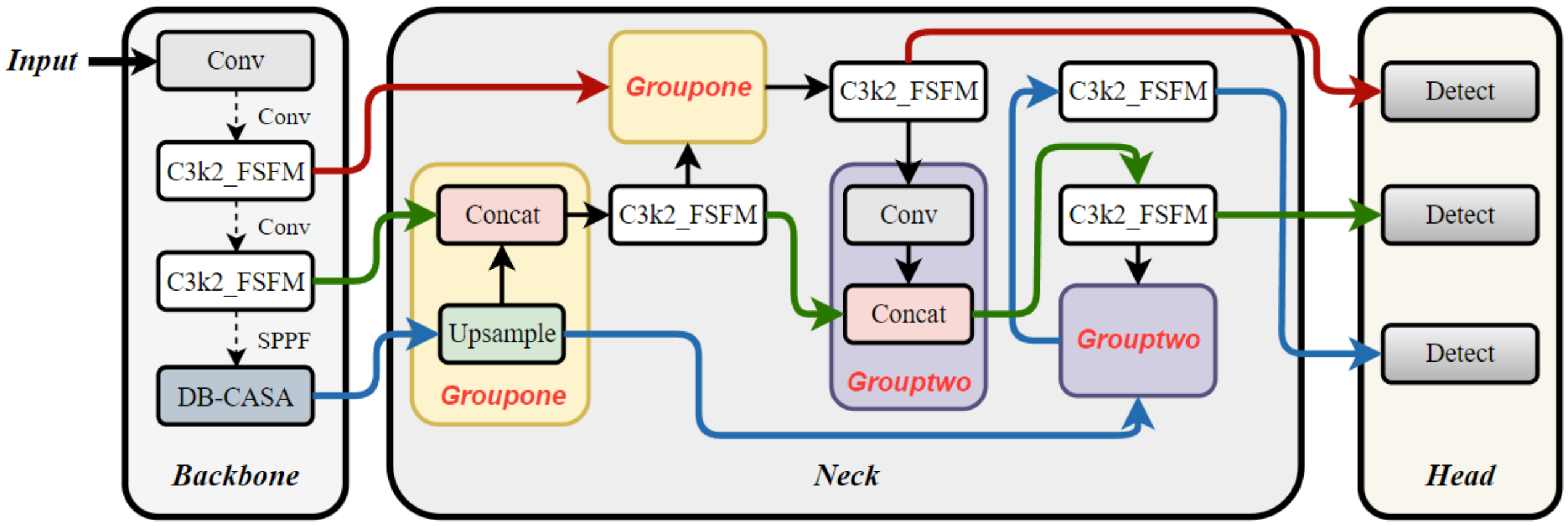

Fig 1. Schematic of the end-to-end framework of the YOLO-MD marine debris detector. The dashed line indicates an aggregation module and a C3K2 module.

## II. PROPOSED METHOD

We propose YOLO-MD, an enhanced YOLO-based framework with improvements in feature interaction, local representation, and loss design. Specifically, the model incorporates a spatial-channel augmented self-attention mechanism to enhance spatial–channel interactions, a shift-based feature exchange mechanism to improve small-object representation, and a dynamic loss function to address sample imbalance. The overall architecture is illustrated in Fig. 1.

### *A. Dual-Branch Convolutional Enhanced Self-Attention*

Small targets in low-resolution and blurred images are often suppressed by global feature aggregation. To address this issue, we propose a Dual-Branch Convolutional Enhanced Self-Attention (DB-CASA) Module, as illustrated in Fig. 2. which integrates self-attention with convolution to jointly model global dependencies and local details.

Specifically, a Token Mixer is designed to decompose input features into two complementary branches: a spatial interaction branch and a channel interaction branch.

The spatial branch enhances local structural information, while the channel branch captures inter-channel dependencies, enabling effective spatial–channel feature interaction. Unlike conventional attention mechanisms that rely on expensive similarity matrix computation, DB-CASA adopts a lightweight design with cross-branch interactions, applies spatial and channel attention to Query and Key separately, and employs additive fusion instead of concatenation, reducing computational cost while maintaining representation capability.

In the spatial interaction branch, convolution is employed to enhance the boundary and texture features of small targets. In detail, the input feature map $x \in \mathbb{R}^{H \times W \times C}$ is first processed by DWConv [14] to capture local context, followed by a 1 × 1 convolution to reduce the channel dimension to 1, generating the spatial attention map $F_s(x) \in \mathbb{R}^{H \times W \times 1}$.

$$F_s(x) = \sigma(Conv_{1\times1}(Conv_{\mathrm{FD,ReLU,BN}}(x))) \odot x \qquad (1)$$

This operation strengthens the spatial discriminability of small targets in the feature map, ensuring they are not obscured during global computation.

In the channel interaction branch, inspired by SENet [15], we combine convolution with global average pooling to integrate information across channels without reducing the channel count, generating a channel attention map $F_c(x)$.

$$F_c(x) = \sigma(Conv_{1\times1}(Pool_{avg}(x))) \odot x \qquad (2)$$

This preserves fine-grained channel-wise information, which is critical for distinguishing small targets with subtle appearance variations.

The Query, Key, and Value are obtained by independent linear transformations, $\mathrm{Q} = \mathrm{W_q x}, \mathrm{K} = \mathrm{W_k x}, \mathrm{V} = \mathrm{W_v x}$. The outputs of the spatial and channel branches are applied to $Q$ and $K$ respectively, then fused via addition. The final output of the DB-CASA module is defined as:

$$OP = L(\mathrm{F_Q}(\mathrm{x}) + \mathrm{F_K}(\mathrm{x})) \odot V \qquad (3)$$

where $L(\cdot)$ denotes the linear transformation for integrating the contextual information. This design enables efficient spatial–channel interaction, improving feature representation while maintaining low computational cost.

### *B. Feature Shift Fusion Module*

To enhance local feature interaction with minimal computational overhead, we propose a Feature Shift Fusion Module (FSFM). Instead of introducing additional convolutional operations, FSFM performs cross-dimensional feature exchange through parameter-free spatial shifts, enabling efficient local context modeling.

The overall structure of FSFM is illustrated in Fig. 3. The module first divides the input feature map into multiple sub-feature groups along the channel dimension:

$$\mathrm{X} = [\mathrm{X_1}, \mathrm{X_2}, \mathrm{X_3}, \mathrm{X_4}] \qquad (4)$$

Spatial shift operations are then applied to different groups in four directions (horizontal and vertical), enabling cross-dimensional feature interaction. Compared with prior unidirectional shifting methods such as Shift-Net [16], this design captures multi-orientation local context without introducing additional parameters. The shifted features are subsequently concatenated along the channel dimension to form the fused representation:

$$\mathrm{X_s} = \mathrm{Concat}\left(\mathrm{shift}_w^{+}(\mathrm{X_1}), \mathrm{shift}_w^{-}(\mathrm{X_2}), \mathrm{shift}_h^{+}(\mathrm{X_3}), \mathrm{shift}_h^{-}(\mathrm{X_4})\right) \qquad (5)$$

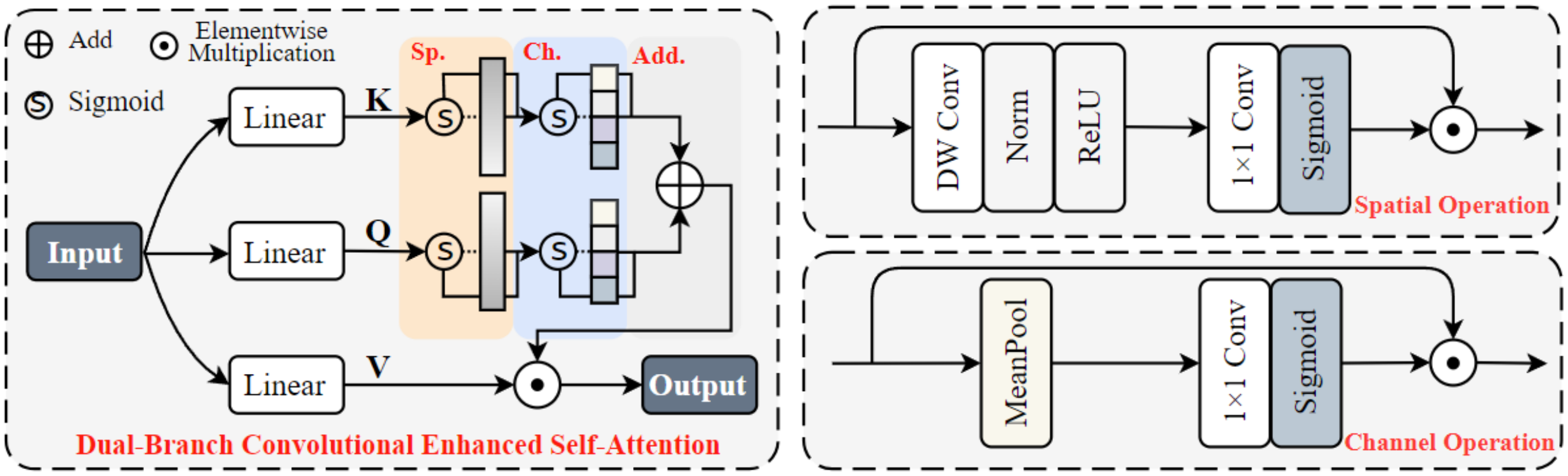

Fig 2. The structure diagram of Dual-Branch Convolutional Enhanced Self-Attention. In this figure, 'Sp.' denotes Spatial Operation and 'Ch.' denotes Channel Operation. The top right section represents operations performed in the spatial domain, while the bottom right section depicts operations conducted in the channel domain.

where $shift^{\pm}(\cdot)$ denotes the shift operation along the specified dimension with step size $s$, and $Concat(\cdot)$ represents channel-wise concatenation.

Integrated within the C3k2 structure, FSFM works with residual connections to facilitate feature propagation and enhance local information preservation.

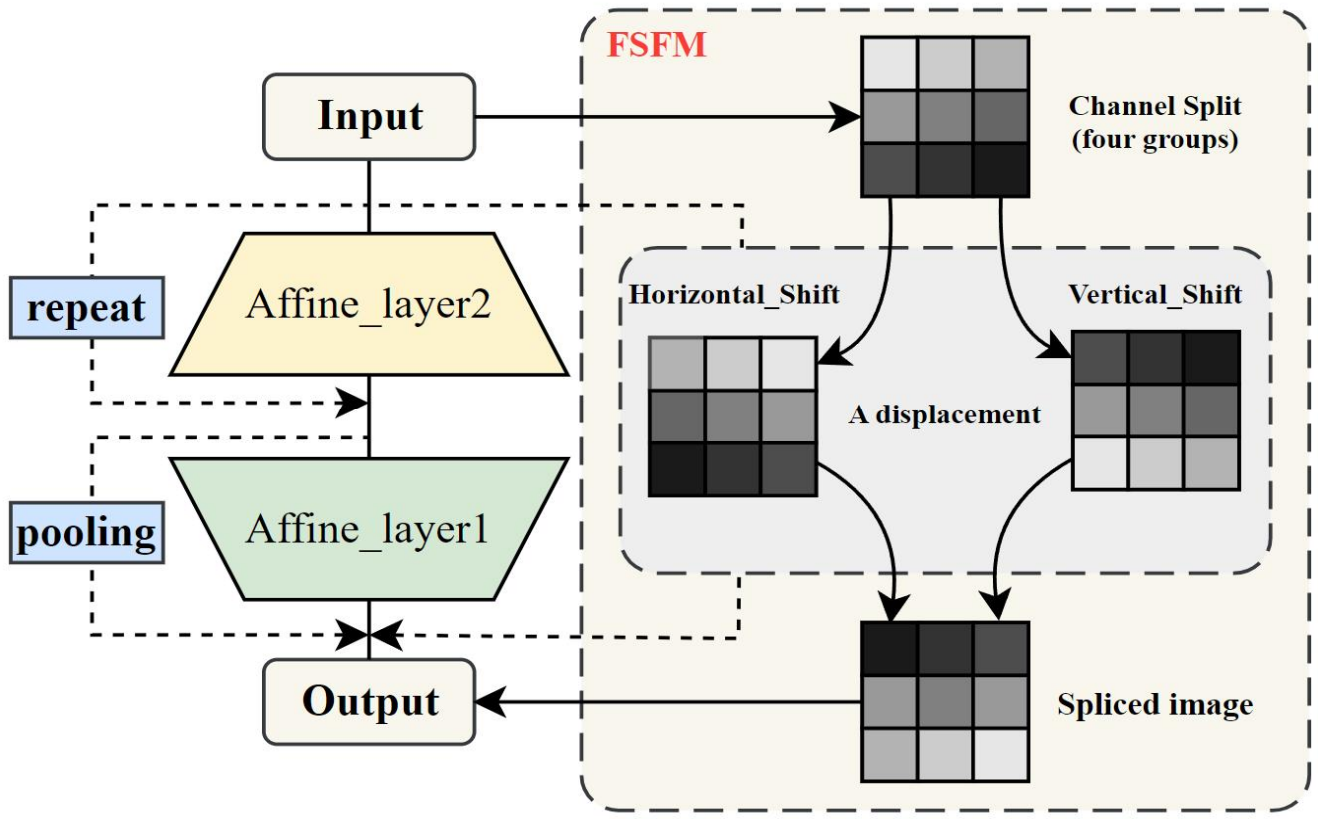

Fig 3. The architecture of Feature Shift Fusion Module-C3k2.

## C. SFG-Loss Function

To address class imbalance and unstable optimization in complex detection scenarios, we propose SFG-Loss, a composite loss that integrates localization accuracy and sample reweighting in a unified framework.

Specifically, we adopt Generalized IoU (GIoU) [17]as the base regression loss to provide stable gradient signals, especially for non-overlapping bounding boxes. To further improve training stability, a dynamic weighting mechanism inspired by Slide Loss is introduced to adaptively emphasize hard samples while suppressing easy ones. The weighting function is defined as:

$$f(x)=\begin{cases}1 & x\le\mu-0.1\\ e^{1-\mu} & \mu-0.1<x<\mu\\ e^{1-x} & x\ge\mu\end{cases} \tag{6}$$

We then multiply this weight by the GIoU loss to obtain Slide-GIoU (SG) [18], which assigns higher importance to hard samples.

In addition, we incorporate Focaler-IoU [19]to constrain extreme cases by truncating excessively easy or hard samples, preventing training instability caused by outliers. Its formulation is given by:

$$SG^{focaler}=\begin{cases}0 & SG<d\\ \dfrac{SG-d}{u-d} & d\le SG\le u\\ 1 & SG>u\end{cases} \tag{7}$$

where d and μ are predetermined thresholds used to define the effective range of IoU. In the experiments, we selected d=0 and u=0.95, which can suppress excessive distortion while ensuring stability.

By jointly considering localization quality, sample difficulty, and training stability, SFG-Loss improves detection performance and convergence efficiency in marine debris detection.

# III. Experiment and Analysis

## A. Experimental Setting

In this study, we employ the open-source UODM dataset [20] from Roboflow for marine debris detection and identification. The dataset contains 5,125 images with a resolution of 640×640 across 15 categories of marine debris. Model performance is evaluated using precision, mAP50, F1-score, and parameter count. All experiments are performed on an NVIDIA RTX 3070 GPU using Python 3.9, CUDA 11.8, and PyTorch 2.0. The model is trained for 200 epochs with an initial learning rate of 0.01, momentum of 0.937, and weight decay of 0.0005.

## B. Comparison Experiments

As shown in Table I, YOLO-MD achieves the best overall performance across Precision, mAP50, and F1-score, outperforming both conventional detectors and recent methods. Compared with the baseline YOLOv11, YOLO-MD improves mAP50 by 2.8% and Precision by 4.0%. Furthermore, YOLO-MD demonstrates consistent improvements over multiple YOLO variants. Notably, our model far surpasses several latest state-of-the-art models, such as YOLO26.

Compared with other architectures detectors, such as SSD and Faster R-CNN, YOLO-MD achieves superior accuracy while significantly reducing the number of parameters, highlighting its efficiency advantage. In addition, compared with the Transformer-based RT-DETR-l, YOLO-MD achieves higher detection accuracy with substantially lower computational cost, indicating a more favorable trade-off between performance and efficiency.

TABLE I
COMPARATIVE EXPERIMENTS ON THE UODM DATASET

| Algorithm | P | mAP50 | F1 | Params/M |
|---|---|---|---|---|
| RT-DETR-l[12] | 0.808 | 0.780 | 0.785 | 32.01 |
| SSD[7] | 0.762 | 0.641 | 0.810 | 25.52 |
| Faster-RCNN[6] | 0.647 | 0.820 | 0.750 | 115.80 |
| YOLOv5[21] | 0.823 | 0.811 | 0.788 | 2.51 |
| YOLOv8[22] | 0.822 | 0.814 | 0.802 | 3.01 |
| YOLOv10[23] | 0.792 | 0.786 | 0.761 | 2.70 |
| YOLOv11[24] | 0.841 | 0.826 | 0.790 | 2.59 |
| YOLOv12[25] | 0.824 | 0.794 | 0.773 | 2.54 |
| YOLOv13[26] | 0.827 | 0.825 | 0.807 | 2.45 |
| YOLO26[27] | 0.758 | 0.765 | 0.744 | **2.38** |
| Ours | **0.875** | **0.849** | **0.822** | 2.78 |

Qualitative results are illustrated in Fig. 4. Compared with DETR-L and YOLOv11, YOLO-MD exhibits stronger capability in detecting small and blurred targets and provides more accurate localization.

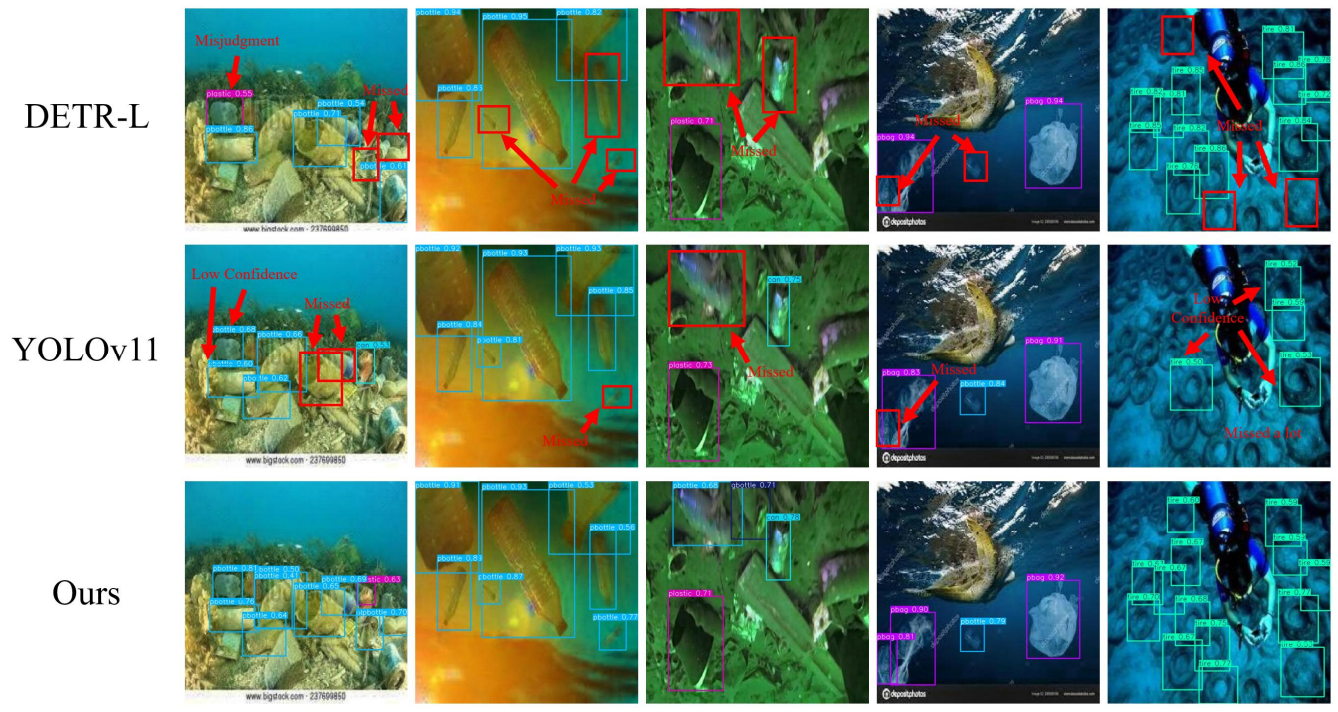

Fig 4. The detect performance comparison on UODM dataset. YOLO-MD effectively detects small targets that are often overlooked by other advanced models, demonstrating superior accuracy in localizing target boundaries.

## C. *Ablation Experiments*

Each module improves performance from a different perspective. Table II reports the ablation results. Individually, DB-CASA improves F1-score by 2.8%, indicating enhanced recall for small targets, while FSFM boosts precision to 0.876, demonstrating its effectiveness in refining localization. Notably, the combination of DB-CASA and FSFM yields a slight drop in precision, suggesting a mild trade-off between global attention and local shifting. This is effectively resolved by SFG-Loss, which restores balance and drives the full model to the optimal mAP50 of 0.849 and F1-score of 0.822.

TABLE II
ABLATION EXPERIMENTS ON THE UODM DATASET

| DB-CASA | FSFM | SFG-Loss | P | mAP50 | F1 | Params/M |
|---|---|---|---|---|---|---|
| | | | 0.841 | 0.826 | 0.790 | 2.59 |
| √ | | | 0.842 | 0.840 | 0.818 | 2.62 |
| | √ | | 0.876 | 0.833 | 0.806 | 2.74 |
| | | √ | 0.828 | 0.827 | 0.802 | 2.59 |
| √ | √ | | 0.821 | 0.835 | 0.809 | 2.78 |
| √ | | √ | 0.845 | 0.835 | 0.813 | 2.62 |
| | √ | √ | 0.850 | 0.818 | 0.789 | 2.75 |
| √ | √ | √ | **0.875** | **0.849** | **0.822** | 2.78 |

Overall, the three modules are complementary, and their integration achieves the best performance with only a marginal increase in parameters, confirming the effectiveness and efficiency of the proposed design.

## D. *Edge Deployment*

In addition to benchmark evaluation, we further validate YOLO-MD in a real-world edge deployment scenario. The system consists of an unmanned surface vehicle equipped with an underwater camera and an embedded edge computing device for on-board image acquisition and real-time detection, eliminating the need for communication with a remote host computer.

Field experiments are conducted in a campus artificial lake, where water conditions vary in turbidity and illumination. These factors make detection more difficult than in controlled datasets. The experimental platform and edge deployment scenario are shown in Fig. 5.

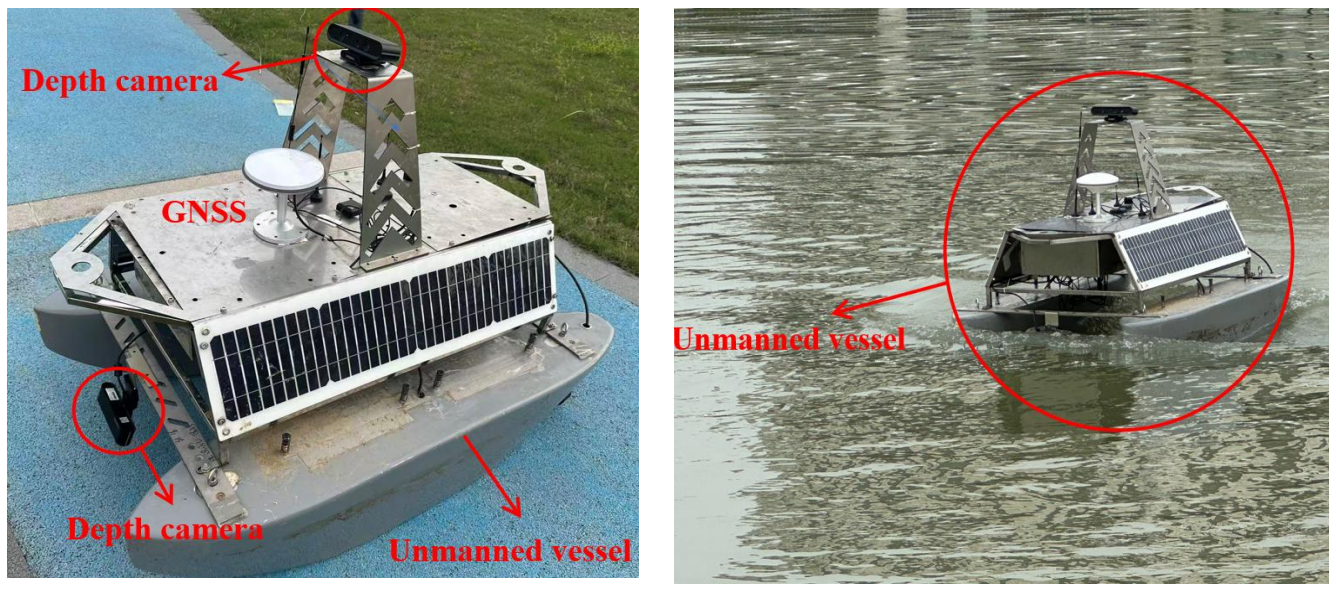

Fig 5. Demonstrations of unmanned surface vessel used for field experiments and on-site experiments.

Representative detection results are shown in Fig. 6. Despite the degraded visual conditions encountered during field deployment, characterized by low illumination, elevated water turbidity, and consequent image blur, YOLO-MD consistently localizes and recognizes multiple types of submerged debris with clear boundaries in real time on the edge device. These results confirm the effectiveness and practical applicability of the proposed model for edge-based robotic perception in challenging underwater environments.

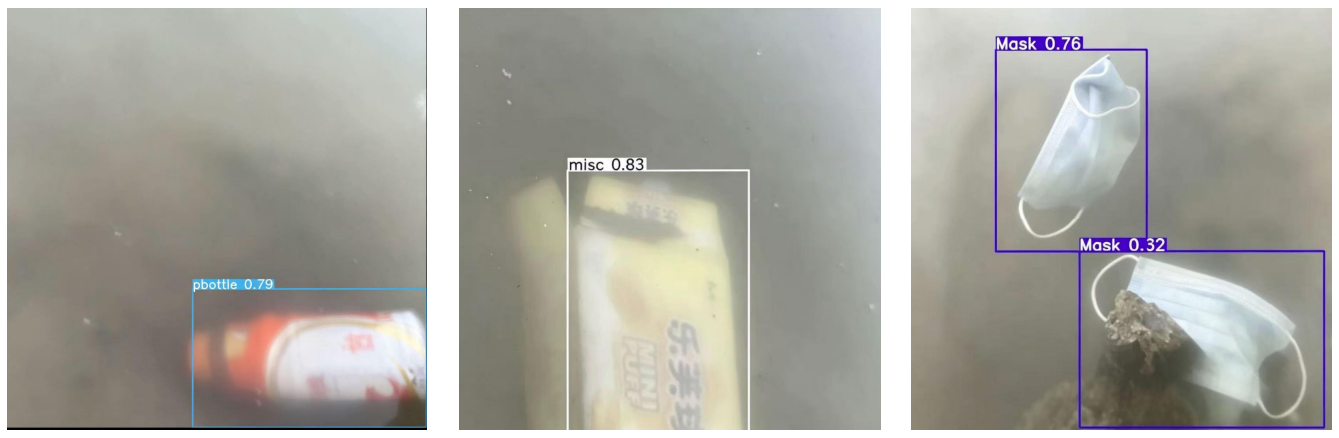

Fig 6. Results of field experiments.

# IV. Conclusion

YOLO-MD is proposed for marine debris detection, achieving improved performance through enhanced feature representation, shift-based local interaction, and an optimized training strategy. Experimental results on the UODM dataset demonstrate that the proposed method outperforms existing detectors in both precision and efficiency. Field deployment on an unmanned surface vehicle further validates its effectiveness in degraded underwater environments. Future work will focus on further research on generalization capability, as well as field experiments targeting small objects in marine environments.